# Efficient Licence Plate Detection By Unique Edge Detection Algorithm and Smarter Interpretation Through IoT.

Tejas K, Ashok Reddy K, Pradeep Reddy D, Rajesh Kumar M, *Senior Member, IEEE*
Department of Electronics and Communication Engineering
VIT University
Vellore, India 632014
Email: tejastk.reddy@gmail.com, mrajeshkumar@vit.ac.in

*Abstract*— **Vehicles play a vital role in modern day transportation systems. Number plate provides a standard means of identification for any vehicle. To serve this purpose, automatic licence plate recognition system was developed. This consisted of four major steps: Pre-processing of obtained image, extraction of licence plate region, segmentation and character recognition. In earlier research, direct application of sobel edge detection algorithm or applying threshold were used as key steps to extract the licence plate region, which do not produce efficient results when captured image is subjected to high intensity of light. The use of morphological operations causes deformity in the characters during segmentation. We propose a novel algorithm to tackle the mentioned issues through a unique edge detection algorithm. It is also a tedious task to create and update the database of required vehicles frequently. This problem is solved by the use of 'Internet of things' where an online database can be created and updated from any module instantly. Also, through IoT we connect all the cameras in a geographical area to one server to create a 'universal eye' which drastically increases the probability of tracing a vehicle over having manual database attached to each camera for identification purpose**

## I.  INTRODUCTION

The exponential growth in the number of vehicles used in cities and towns is leading to increase in various crimes such as theft of vehicles, violating traffic rules, hit and run accidents and trespassing into unauthorised areas etc. Therefore adapting automatic licence plate recognition system is much necessary. Digital image processing techniques were used to interpret digitized images in order to extract meaningful knowledge from it. These techniques were broadly classified into four stages: Pre-processing of the captured vehicle licence plate image, detection and extraction of licence plate region, segmentation of characters from the extracted plate using morphological operations and character recognition. In earlier research, the extraction of licence plate region was either done through direct application of sobel edge detector followed by linear dilation and filling connected components with holes or the image was converted into a binary image by applying threshold at a certain value. These methods do not produce efficient results when captured licence plate image is subjected to high intensity of light or when the vehicle and licence plate are of similar colours. This problem also exists due to variable factors such as distance, weather and camera quality. In this paper we discuss a novel algorithm to solve these issues to an extent. In the segmentation process morphological operations such as dilation and erosion are used which causes deformity in the characters. The original font style and size are manipulated, which makes it hard for character recognition. A unique algorithm is developed for the segmentation of characters without morphological operations for efficient results in this paper.

Internet of Things (IoT) is a technology where an embedded system is connected to the cloud for easier and smarter analysis. The hardware part of embedded system collects the data and software part process the data to produce use full information. This information is passed on to the cloud where it can be interpreted. The use of IoT technology has been increasing exponentially in the recent years. Its drastic growth is due its ability to make things smart and make it easier to collect and interpret large sets of data through cloud computing. The combination of automatic licence plate recognition system with IoT can produce much efficient results. Here the camera acts as the embedded hardware which collects the data through digitized vehicle images and MATLAB 2016 software is used to process the image to obtain the licence plate number, which is then transferred to cloud through internet for further applications.

All surveillance cameras as well as traffic control cameras can be connected to a cloud which consists of database of required vehicles. By doing this, all the cameras in an area act as a single system where each camera acts as an 'eye' searching for the vehicles listed in the database uploaded by the administrator. In this way a 'universal eye' could be created making it hard for the accused to hide the vehicle and also saves time on searching for these vehicles.

## II. LITERATURE REVIEW

The Automatic Number Plate Recognition (ANPR) was invented in 1976 at the Police Scientific Development Branch in the UK. Sang Kyoon Kim et al [1] proposed a method using a distributed genetic algorithm to overcome the difficulties dealing with degraded number plates. Clemens Arth et al [2] presented a real-time license plate detection and recognition system where image fragments are taken from real-time video and processed. Serkan Ozbay et al [3] proposed a method to add a simple step before character segmentation which is more efficient to remove noise and unwanted spots. Humayun Karim Sulehria et al [4] proposed an algorithm that allows the recognition of vehicles' number plates using hybrid morphological techniques including hat transformations with morphological gradients and neural networks. Muhammad Tahir et al [5] proposed a method for automatic number plate recognition which uses yellow search algorithm to extract the likelihood ROI in an image especially for yellow licence plate recognition. Ankush Roy et al [6] proposed a method to overcome the difficult to correctly identify the non-standard number plate characters, by using a pixel based segmentation algorithm of the alphanumeric characters in the license plate. Arulmozhi K et al [7] applied skew correction technique is applied for accurate character segmentation followed by character recognition Centroid based Hough Transform technique is presented for skew correction of license plate. Amninder Kaur et al [8] performed pre-processing and number plate localization by using Ostu's methods and feature based localization methods. Roy et al [9] proposed a method where image filled with holes is used for licence plate segmentation, removing all the connecting edges and applied a threshold of 1000 pixels. Shan Du et al [10] presented all the methods which are being used till then and have also listed pros and cons of each method in plate segmentation. Gee-Sern Hsu et al [11] used Edge clustering formula for solving plate detection for the first time. It is also a novel application of the maximally stable extreme region (MSER) detector to character segmentation. Norizam Sulaiman et al [12] proposed a combination of image processing techniques and OCR to obtain the accurate vehicle plate recognition for vehicles in Malaysia. Bhat et al [13] have also used licence plate segmentation through filling the binary image with holes but this system had limitations on size of vehicle. Sarbjit Kaur et al [14] applied bilateral filter during pre-processing for the image and remaining method is same as [9] and [13]. Chi-Hung Chuang et al [15] proposed an approach to overcome the problems like low resolution, long distance image and blurred image with super resolution, by LBP with the concept of fuzzy. Jitendra Sharma et al [16] concentrated on the improvement of the recognition rate and recognition time for recognition of the number and the characters of the vehicle license plate. Ayman Rabee et al [17] proposed an algorithm where SVM was applied at last to construct a classifier to categorise the inputs. Priyanka Prabhakar et al [18] proposed an algorithm which uses Hough transform for licence plate segmentation. Raskar R et al [19] used median filter to remove the noise and later followed sobel edge detection algorithm. Silvestre Garcia-Sanchez et al [20] proposed a method for identifying the position, length and width of the licence plate that contains the characters in an image, when the plate's location has not been accurate. Sanchay Dewan et al [21] proposed an ant colony based number plate extraction method which serves better results in edge detection while applying image segmentation. Ayush Kapoor et al [22] presented a method were IoT and image processing are combined in the field of agriculture the input data is tested comparing with the pre-defined data base of the images which are already fed, this gives the information of character of input.

## III. PROPOSED METHOD

### 1.1. Digital Image Processing techniques:

Image processing techniques are used to convert the raw data obtained by the cameras into useful information. This algorithm can be broadly classified into four stages which can be discussed as follows.

### 1.1.1. Pre-processing

The video or image of the vehicle is obtained from the camera, whose vehicle number is to be identified as illustrated in Fig.2(b). If it was a video, the video is converted into frames and these frames are further used for vehicle number identification. This image or frame is then converted from RGB scale to grey scale. A median filter is applied on the obtained grey scale image in order to remove different kinds of noises present in the image. The median filter also concentrates on the high frequency areas of the image as shown in Fig.1, which helps to obtain better edge detection results in the later part of the algorithm.

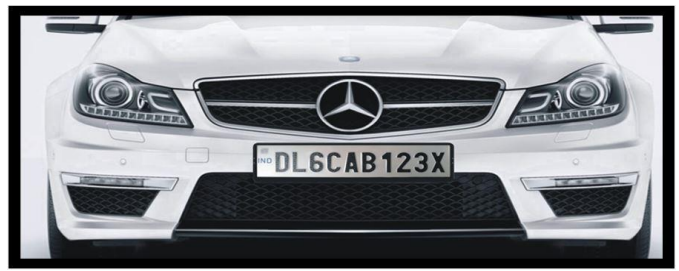

Fig.1. Image after applying median filter

*1.1.2. Extraction of licence plate region*

Here our aim is to localise the position of the licence plate region and extract a sub-image plot with only the number plate for further analysis as described in Fig2.(c).The image obtained after processing through a median filter is now processed through an averaging filter with structuring element of size [20 x 20], which returns the centre part of the correlation value without zero padding at the edges. As a result of which a blurred image is obtained. Now, the blurred image is subtracted from the original grey scale image to obtain intensity difference image. This is because, when an image is blurred by using an average filter the high frequency pixels in the image tend to equalise their pixel value with their surroundings. Thus surrounding pixels acquire a higher value.

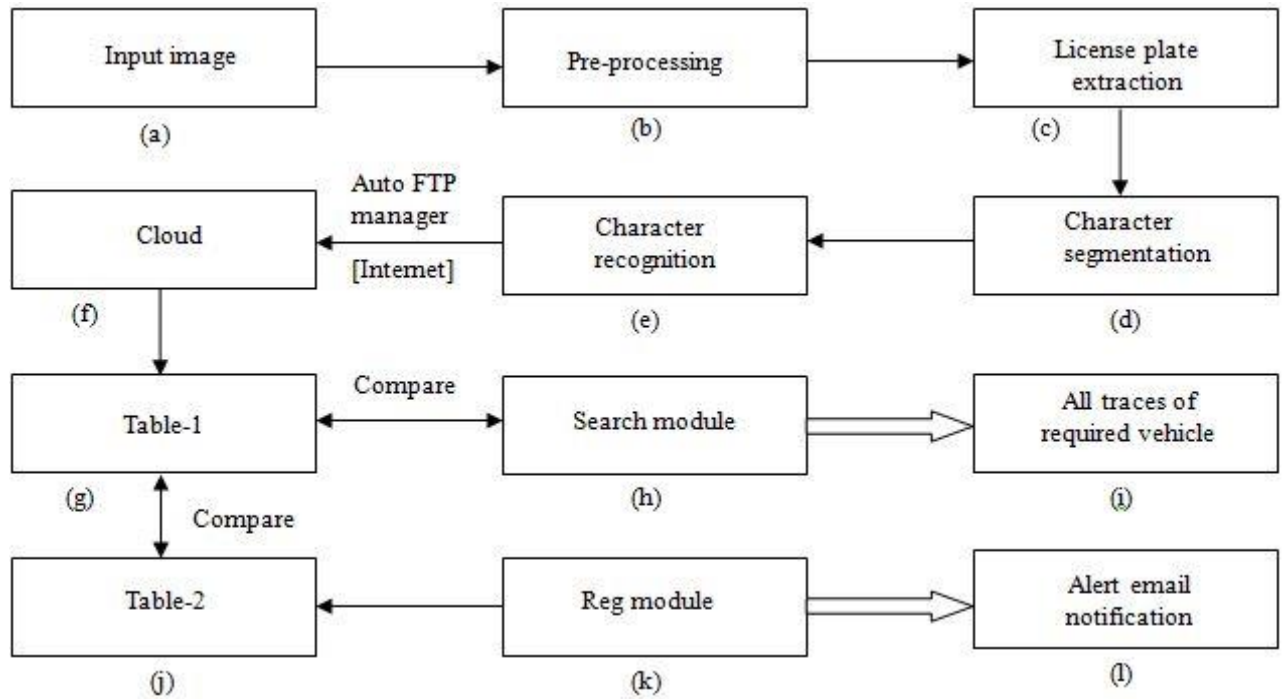

Fig.2. Flow chart of the proposed algorithm

When this blurred image is subtracted with original grey scale image, the extra acquired values remain and others are deleted. Resulting in an image as shown in Fig.3.(a). Threshold this image at a very low value, to create a binary image. For efficient results threshold at a value of 0.03, where all pixels above this value tend to become unity and others tend to become zero. All the components touching the boundary of the obtained binary image are then deleted.

Sobel edge filter is applied on the obtained binary image in order to obtain accurate boundaries of binary objects in the figure. Scan for all the connected components in the figure and fill them with holes as shown in Fig.3. (b). Search for a rectangular area filled with holes in the image which is probably in the size of the licence plate. This rectangular region of the licence plate is now extracted into a sub image figure. Multiply the extracted licence plate region with the initial grey scale image, in order to isolate the licence plate from the original image of the vehicle as shown in Fig. 3. (c)

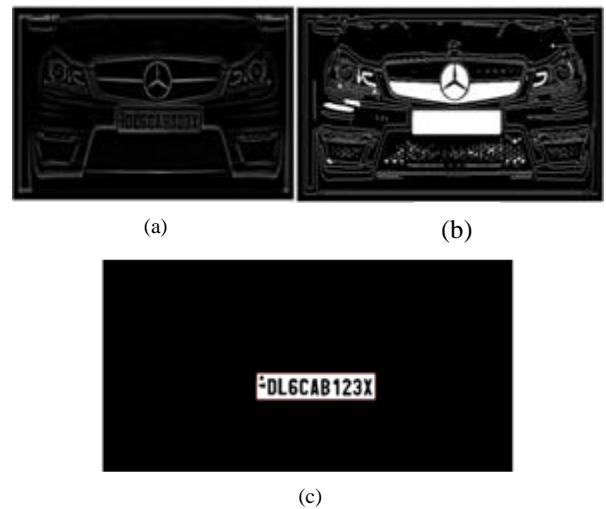

Fig.3. (a) subtracted image, (b) sobel edge detected image
(c) Licence plate extracted image.

### 1.1.1. Character Segmentation

Threshold this obtained licence plate at a low value such as 0.01 to convert the grey-scale licence plate image into a binary image

The filled regions in the image are now labelled with distinct indices and the bounding box of four sides is applied on the image to mark all the filled regions of the image. Initially to remove the extra zero padded regions in the image, bounding box algorithm is applied to obtain the rectangle of maximum size. The licence plate is extracted in this step which is resized to a fixed size for easier character recognition in the further part. Consider the fixed size to be [175x730]. The obtained licence plate is now complemented and the boundaries are cleared to erase the licence plate border or unnecessary regions leaving only the characters in the image. The regions with less than 1000 pixels and more than 8000 pixels is deleted in order to remove the rusted part, screw holes or broken regions of the number plate (if any) and results in an image with characters as shown in Fig.4.(a). The remaining characters in the image are now individually extracted by applying bounding box algorithm again as shown in Fig.4.(b).

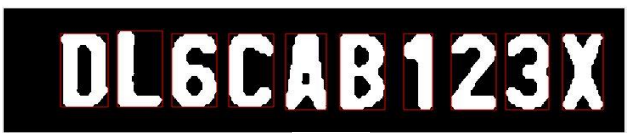

(a)

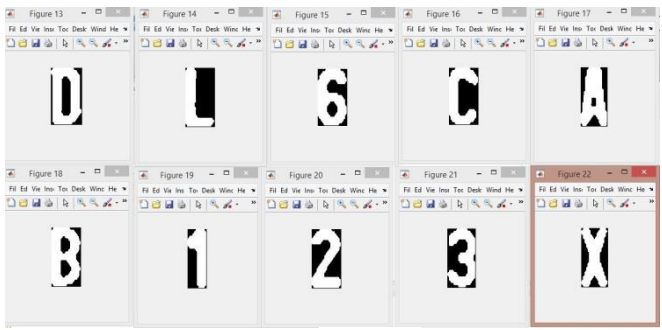

(b)

Fig.4. (a) image with characters, unnecessary regions removed. (b) Individually segmented characters.

### 1.1.2. Character recognition

The next step is identifying the information in the image and decoding it for further interpretation as seen in Fig.2.(e). Each character from the licence plate is extracted by applying bounding box algorithm and these characters are compared against the alphanumeric database which uses template matching. The extracted character is compared with the template images in all possible positions.

A large database of alpha numeric characters can be taken with each character having several font styles and sizes. Neural logic's can be applied to train the database in order to identify different font styles. New font styles can be regularly added during the process of character segmentation. In this way, the algorithm becomes smarter making it artificially intelligent. The final output is written in a notepad file as shown in Fig.5.

### 1.2. Internet of Things technology:

Raw data is converted into information by the image processing techniques. This information is interpreted using IoT technology. The information produced in the notepad file is then transferred to the cloud for further use as illustrated in Fig.2.(f). To interpret the uploaded information we need an internet application and hence we created a tentative website *www.searchyourcar.ml*

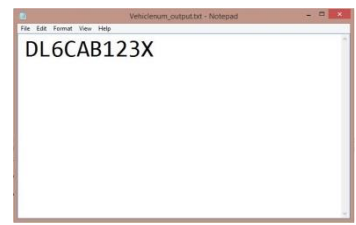

Fig.5. Character recognition output

.
This website was hosted from *www.hostinger.in*. The database for this web application is phpMyAdmin and the online transfer's takes place using mySQL queries. The front end user interface part of the web application in the website is coded using Hypertext Markup Language (HTML) and Cascading Style Sheets (CSS). The back end database for the web application is phpMyAdmin and is maintained using mySQL queries. The connection between front end and the back end of the web application is coded using PHP.

The output from the MATLAB software is written and stored in a notepad file. The information in the notepad file is later read and transferred to the cloud using Auto File Transfer Protocol (FTP) Manager once in every 10 seconds which is later used for interpretation. This information can be accessed and interpreted using two different modules as discussed further.

### 1.2.1. Search Module

The main aim of search module is to know the vehicle number in interest of the administrator and input the number from user interface of the website. This vehicle number is then searched in the database and all its traces in different locations and time can be accessed by the authorised administrator. The user interface of search module is shown in Fig.6.

In the server side we have two different tables for managing the data taken from the front end of the database. Table - 1 contains the attributes such as vehicle number, location (latitude and longitude), time and date. Table - 2 is used to manage the register module as shown in Fig.2.(h),(k). The search module algorithm is discussed in detail in Algorithm.1.

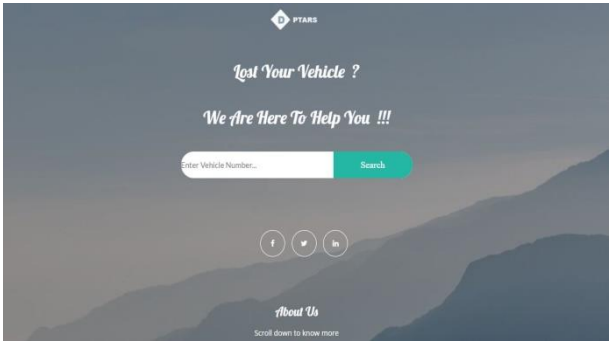

**Fig.6.** Search module in the web application

---

**Algorithm 1:**

Input: vehicle number from user interface.

Output: Trace of that vehicle at different locations in different situations.

1. Establishing the connection with server by matching username, password and database name by using connection object in PHP using mySQL queries.
2. Input is taken from the interface and compared with the data in the table.
3. SELECT number, location, time FROM 'Table -1' WHERE number='input from user'.
4. Data from user is transferred to server using Auto FTP manager in Table-1.
5. Using Jason decoding algorithm, location is retrieved by using IP address from the source 'http://freegeoip.net'.
6. Local time zone is set according to locality in PHP time zone. This time is retrieved to store the vehicle traced time in Table -1.
7. Using Insert MySQL query data is inserted to Table-1.

---

### 1.1.1. Register module

In case we could not trace a vehicle through search module immediately, then we could use register module. Register module obtains data about the required vehicles through attributes such as vehicle number, user email id, mobile number and the details of the required vehicle. Through this module, the required vehicle number is compared with every input number uploaded from the embedded system. If a match is found, an alert email is sent to the registered email id. Register module algorithm can be discussed in detail as Algorithm 2. The front end user interface of the register module is as shown in Fig.7.

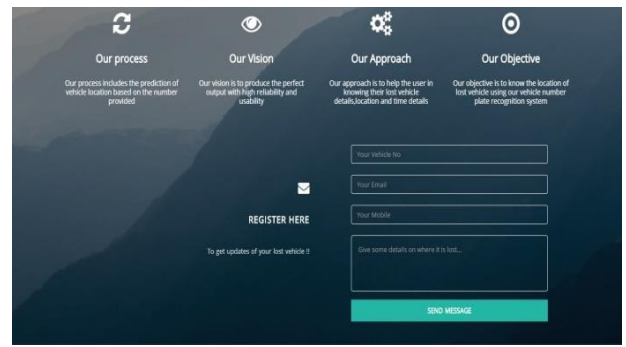

Fig.7. Register module in the web application

---

**Algorithm 2:**

Input: Vehicle number, email id.
Output: An alert email notification if the vehicle is traced.

1. The data such as vehicle number, email id, mobile number and details of the required vehicle are collected from the admin using HTML form in the front end of the web application.
2. Using PHP, the connection is established between the front end and the back end of the website. Here, the Table -2 acts as the back end of the database.
3. The attributes collected in step 1 are stored in table - 2 by using INSERT mySQL query.
4. The vehicle numbers transferred through FTP manager to Table -1 are automatically compared with the vehicle in table -2. If a match is found then the email id is retrieved from table -2.
5. Using PHP mailer, an alert mail is immediately sent to the email id retrieved from Table-2.

---

## IV. EXPERIMENTAL RESULTS

Table I: Extracting license plate region by converting input image into binary image

| Recent research | Input image | Existing method | Proposed method |
|---|---|---|---|
| (a) S Kaur et al [14] | 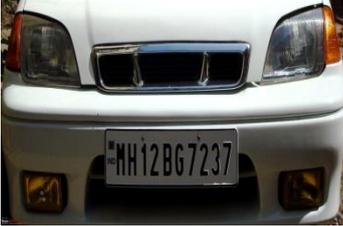 | 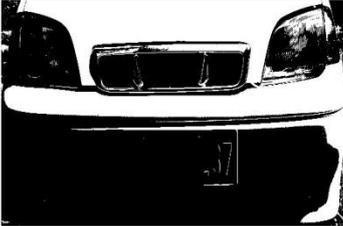 | 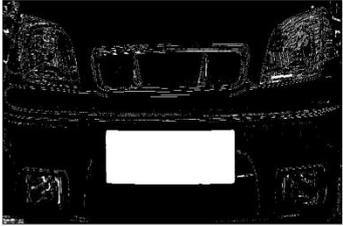 |
| (b) J A Khan [23] R R Raskar [19] | 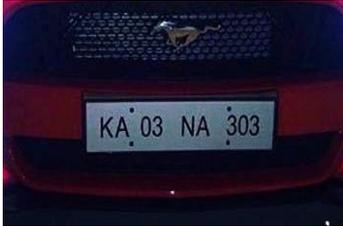 | 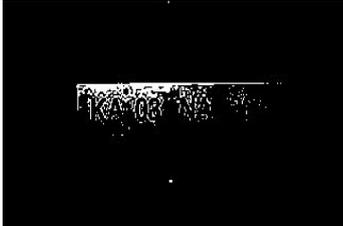 | 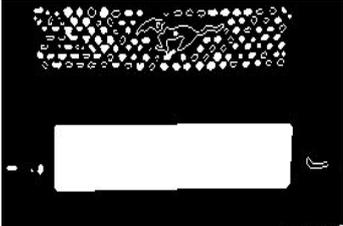 |

Table II: Extracting license plate region by applying sobel edge mask on input image

| Recent research | Input image | Existing method | Proposed method |
|---|---|---|---|
| (a) Saurav Roy et al[9] Bhat [13] | 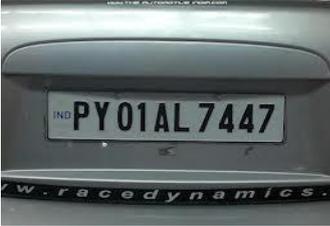 | 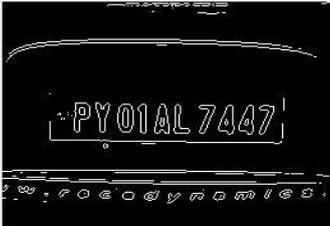 | 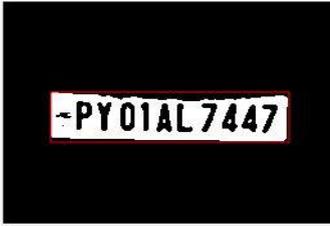 |
| (b) K M Babu et al[24] | 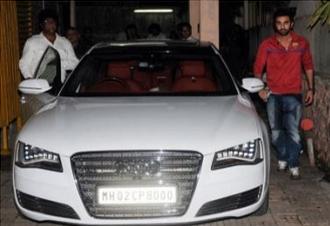 | 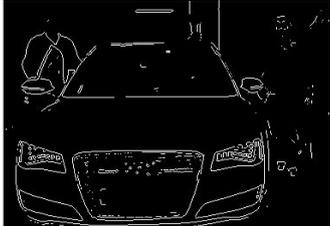 | 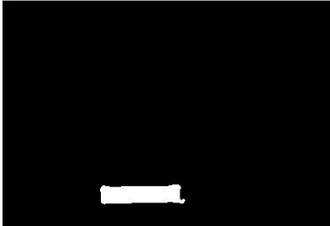 |

Table III: Character Segmentation

| Recent research | Input image | Existing method | Proposed method |
|---|---|---|---|
| (a) Dewan S et al[21] | 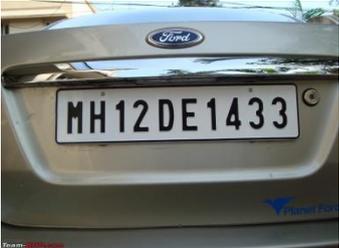 | 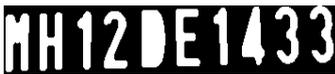 | 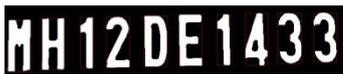 |

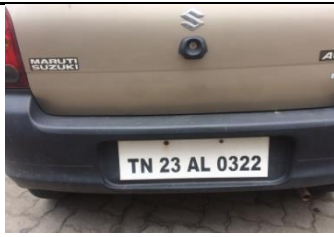

(b)
A Agarwal [25]

## 4.1. Digital image processing

The input image is taken from an embedded camera and this image is processed through digital image processing techniques, which were carried out in MATLAB 2016 software. Unique edge detecting algorithm was used to improve the efficiency of extracting the licence plate region. A distinct algorithm for segmenting the characters in the number plate was proposed in the paper which avoided producing deformity in the characters.

In Table - I it is observed that direct conversion of the input grey scale image to a binary image by applying threshold at 127 pixel values in order to detect the edges of the licence plate do not produce efficient results when the light intensity is not uniform throughout the picture. We can see that in Table - I (a) The intensity of light is more towards right side of the picture as compared to the left and in Table - I (b) The intensity of light is non-uniform throughout the picture. The proposed algorithm improves the efficiency in detecting and extracting licence plate region for images that are subjected to non-uniform intensity of light.

In Table - II we can see that, applying sobel edge mask directly on input grey scale image in order to obtain the licence plate region does not produce efficient results when the input image is subjected to high intensity of light, when the shadow of the vehicle is on the licence plate or when the licence plate tends to reflect light. In Table - II (a) we can see that the proposed algorithm could extract the licence plate region more efficiently as compared to the existing method. In Table - II (b) we can see that the licence plate is reflecting light, which hinders efficient licence plate detection and extraction.

In Table-III it can be observed that, the use of morphological operations such as dilation and erosion produces deformity in the characters. This deformity further hinders the process of character recognition, where characters such as 'O', 'D', '0' and 'B' are falsely recognised among each other when filled with holes due to irregular dilation. This deformity can be eliminated by the use of distinct algorithm proposed in this paper where bounding box method is used twice in order to segment the characters.

## 4.2. Smarter recognition using IoT technology

After processing the input image through MATLAB, the vehicle number is written and saved in Notepad file. This information written in the file is constantly cleared and new information is uploaded to the cloud from the notepad with the help of auto FTP manager. This uploaded information is stored in Table-1 as shown in Fig.8. (a). Now in order to trace a vehicle, we can either use search module to check if the trace of the vehicle is already exists in recorded database or we can use register module to register a required vehicle number with an email id so that an alert email notification is sent as soon as the cloud traces the vehicle number. If a trace if found through search module the administrator can obtain the location, date and time where the vehicle was traced as shown in Fig.8. (c). The details of registered numbers are stored in table-2 as shown in Fig.8. (d). the values in Table – 2 are constantly compared with vehicle numbers in table – 1 and if the match is found, an alert email is sent to the registered email id as shown in Fig.7. (b) from which the administrator can obtain the location where the vehicle was traced along with time and date.

(a)

(b)

|                    | LAST TRACED AT !                          |                     |
|--------------------|-------------------------------------------|---------------------|
| NUMBER             | LAT \|\| LONG & LOCATION                  | DATE & TIME         |
| TN23CB0624         | 12.9333 \| 79.1333 \| Vellore \| TN \| IN | 2017-05-28 22:20:05 |

SEARCH YOUR VEHICLE LOCATION FROM HERE -->

(c)

| vehicle     | email                    | mobile     | details                            |
|-------------|--------------------------|------------|------------------------------------|
| TN29AE5417  | pradeepreddy0003@gmail.com | 8688114776 | I lost my vehicle near banjala palace. |
| TN23CB0624  | pradeepreddy0003@gmail.com | 9994370499 | I lost my vehicle in vellore.      |

(d)

Fig.8. Results of proposed algorithm where (a) Table-1, (b) register module results, (c) Search module results and (d) Table-2

**Table 3: Results**

| Units         | Number of accuracy | Percentage of accuracy |
|---------------|--------------------|------------------------|
| Extraction    | 93/95              | 97.89%                 |
| Segmentation  | 94/95              | 98.947%                |
| Recognition   | 92/95              | 96.842%                |

A total of 95 images were processed through MATLAB 2016 software. These images were taken in different illumination conditions. Images were obtained from various cameras from 5 MP to 13 MP clarity. Distance varied from 1 meter to 20 meters. Images of vehicles in various weather conditions were processed. The results after processing are presented in Table.3.

V. CONCLUSION AND FUTURE WORK

In this paper we combined digital image processing techniques with IoT technology to create a much efficient automatic licence plate detection and recognition system. In digital image processing techniques a unique edge detection algorithm was proposed for better licence plate extraction. Bounding box was used to extract the characters from the licence plate which does not produce any distortion in the segmented characters. This processed information was later sent to cloud for further interpretation. This IoT based licence plate recognition makes it easier to update the required database and also makes it easier to trace the required vehicles if all the cameras in a geographical area are connected to one server. As a part of future work, a genetic algorithm can be introduced which can select the best frame in a video in order to make the process faster. The efficiency of the web application can be integrated as per the administrator's requirement.